%% file: main.tex
\documentclass[sigconf]{acmart}

\AtBeginDocument{%
  }

\newcommand{\bheading}[1]{\vspace*{.5em}\noindent{\textbf{#1.}}}
\newcommand{\blue}[1]{\textcolor{black}{#1}}

\copyrightyear{2023}  
\acmYear{2023}  
\setcopyright{rightsretained}  
\acmConference[CHI '23]{Proceedings of the 2023 CHI Conference on Human Factors in Computing Systems}{April 23--28, 2023}{Hamburg, Germany}
\acmBooktitle{Proceedings of the 2023 CHI Conference on Human Factors in Computing Systems (CHI '23), April 23--28, 2023, Hamburg, Germany}\acmDOI{10.1145/3544548.3580903}
\acmISBN{978-1-4503-9421-5/23/04}

\usepackage[english]{babel} 
\usepackage[utf8]{inputenc}
\usepackage{blindtext}

\begin{document}



\title[Examining AI/ML Practitioners' Challenges]{``It is currently hodgepodge'': Examining AI/ML Practitioners' Challenges during Co-production of Responsible AI Values}






\author[R.Varanasi]{Rama Adithya Varanasi}
\affiliation{
 \department{Information Science}
 \institution{Cornell University}
 \city{New York}
 \state{NY}
 \country{USA}
}
\orcid{0000-0003-4485-6663}
\email{rv288@cornell.edu}

\author[N.Goyal]{Nitesh Goyal}
\affiliation{
 \department{Google Research}
 \institution{Google}
 \city{New York}
 \state{NY}
 \country{USA}
}
\email{teshgoyal@acm.org }

\renewcommand{\shortauthors}{Varanasi et al.}

\begin{abstract}
Recently, the AI/ML research community has indicated an urgent need to establish Responsible AI (RAI) values and practices as part of the AI/ML lifecycle. Several organizations and communities are responding to this call by sharing RAI guidelines. However, there are gaps in awareness, deliberation, and execution of such practices for multi-disciplinary ML practitioners. This work contributes to the discussion by unpacking co-production challenges faced by practitioners as they align their RAI values. We interviewed 23 individuals, across 10 organizations, tasked to ship AI/ML based products while upholding RAI norms and found that both top-down and bottom-up institutional structures create burden for different roles preventing them from upholding RAI values, a challenge that is further exacerbated when executing conflicted values. We share multiple value levers used as strategies by the practitioners to resolve their challenges. We end our paper with recommendations for inclusive and equitable RAI value-practices, creating supportive organizational structures and opportunities to further aid practitioners.


\end{abstract}

\ccsdesc[500]{Human-centered computing~Empirical studies in HCI}


\keywords{Responsible AI, RAI, ethical AI, value levers, co-production, collaboration, XAI, FAT, fairness, transparency, accountability, explainability}

\maketitle
\input{1-introduction}
\input{2-literature}

\input{3-methods}

\input{5-1-structures}

\input{5-2-discourse}

\input{5-3-representation}

\input{6-discussion}
\input{7-conclusion}
\bibliographystyle{ACM-Reference-Format}
\bibliography{citations}

\end{document}

%% file: 1-introduction.tex
\section{Introduction}

In November 2021, the UN Educational, Scientific, and Cultural Organization (UNESCO) signed a historic agreement outlining  shared values needed to ensure the development of Responsible Artificial Intelligence (RAI) \cite{unesco2021}. RAI is an umbrella term that comprises different human values, principles, and actions to develop AI ethically and responsibly \cite{Ghallab_2019, Dignum_2021, asken2019therole, rakova2021}. Through UNESCO's agreement, for the first time, 193 countries have standardized recommendations on the ethics of AI. While unprecedented, this agreement is just one of several efforts providing recommendations on different RAI values to be implemented within AI/ML systems \cite{eu2019, Gonzalez_2020, dod2021}. 

In response, several industry organizations have begun to implement the recommendations, creating cross-functional RAI institutional structures and activities that enable practitioners to engage with the RAI values. For instance, several big-tech companies are implementing common RAI values, such as Fairness, Transparency, Accountability, and Privacy, as part of their RAI initiatives \cite{birhane2022, microsoft2022, IBM2022, Fb2021, google2022}. However, such RAI values have minimal overlap with the values prescribed by UNESCO's framework that promotes non-maleficence, diversity, inclusion, and harmony \cite{unesco2020agreement}. Scholars have attributed the lack of overlap to different business and institutional contexts involved in developing AI/ML \footnote {hereafter, we use AI/ML systems to refer to both AI products and ML models} systems \cite{Haubermann_Lutge_2022, jakesh2022how}. Subsequently, it is essential to understand these contexts by engaging with practitioners across multiple roles who come together to co-produce and \textit{enact} such RAI values. Co-production is an iterative process through which organizations produce collective knowledge \cite{jasanoff2004idiom}. During co-production, individual practitioners may hold certain values (e.g., social justice), yet their teams might prioritize other values. \citet{rakova2021} hints at potential challenges that can arise due to such mismatches in RAI values. Our study builds on this critical gap by giving a detailed analysis of those \textit{challenges} and \textit{strategies} (if any) devised to overcome such strains as practitioners co-produce AI/ML systems.

We interviewed 23 practitioners across a variety of roles to understand their RAI value practices and challenges. Our findings show that institutional structures around RAI value co-production contributed to key challenges for the practitioners. We also discovered multiple tensions that arose between roles and organizations during prioritization, deliberation, and implementation.
Interestingly, we also observed development of ten different RAI value levers \cite{Shilton2013, shilton2018}. These are creative activities meant to engage individuals in value conversations that help reduce value tensions. In the remainder of the paper, we first discuss related work about collective values in Responsible AI from an HCI perspective and outline our research questions. We then present the research methodology and results of the study. We conclude with a discussion of our contributions in improving RAI co-production practices of AI practitioners. Overall, this work makes several contributions. First, we describe the experiences and the organizational environment within which AI/ML practitioners co-produce RAI values. Second, we illustrate multiple challenges faced by AI practitioners owing to different organizational structures, resulting in several tensions in co-production. Third, we unpack ten RAI value levers as strategies to overcome challenges and map them on to the RAI values. Lastly, we provide essential strategies at the different levels (individual, and organizational) to better facilitate and sustain RAI value co-production.

%% file: 2-literature.tex
\section{Related Work}

\subsection{The Field of Responsible AI}
In the last decade, Responsible AI (RAI) has grown into an overarching field that aims to make AI/ML more accountable to its outcomes \cite{asken2019therole, Dignum_2019, abrassart_2017}. One of the field's roots lies in Ethical AI, where critical engagement with ethical values in the otherwise traditional AI/ML field have been encouraged \cite{khan1995, Gips1994,ashurst2022, Robbins2019}. Example studies include engagement with core ethical values to provide nuance in technical AI/ML discourse \cite{Harris_Anthis_2021a}, translating ethical values into implementation scenarios \cite{Gips1994,hooker2018}, and AI/ML guidelines \cite{Vakkuri_Abrahamsson_2018, Harris_Anthis_2021a,ethical2019,Hagendorff_2020,Jobin_Ienca_Vayena_2019}, and seminal studies that brought critical ethical problems to the forefront \cite{peter2019, yu2018, Mittelstadt_2019, Awad_2018}.


RAI has also drawn its inspiration from AI for Social Good (AI4SG \cite{ryan2020}) research to study human values more broadly,  going beyond ``ethical'' values. AI4SG helped the RAI field to translate such values embedded in AI/ML systems into positive community outcomes \cite{Berendt_2019} by eliciting specific values (e.g., solidarity \cite{Floridi_2018}), developing methods (e.g., capabilities approach \cite{bondi2021}), and producing representations (e.g., explanations \cite{chinasa2021}) that strongly align with community goals (e.g., the UN Sustainable Development Goals  \cite{Ghallab_2019}). For example, studies have explicitly engaged with underserved communities to examine the impact of the embedded values within AI/ML systems in their lives (e.g., in agriculture \cite{madaan2019}, health \cite{Azra2021,  wang2021}, and education \cite{cannanure2020} domains).
\blue{More recent studies have shed light on how certain practitioners' (e.g., data and crowd workers) practices, contributions, and values are often ignored while developing the AI/ML systems \cite{sambasivan2022, thakkar2022, varanasi2022}. Some have looked at different values (e.g., fairness) that are often ignored by the discriminatory algorithms \cite{ hampton2021, Kalluri_2020, Chouldechova2018}. More recent work at the intersection of these two by \citet{goyal2022your} has also highlighted the impact of data workers of marginalized communities on the AI/ML algorithms to highlight complexity when building for RAI values like equity. }

Lastly, RAI has also drawn motivation from recent movements associated with specific value(s). One such movement is around the value of explainability (or \textit{explainable AI}) that arose from the need to make AI/ML systems more accountable and trustworthy \cite{doran2017, Barredo_2020, ehsan2021}. 
A similar movement within the RAI's purview focused on FATE values (Fairness, Accountability, Transparency, and Ethics/Explainability) \cite{Shin_Park_2019, Licht_2020, kroll165accountable, Mitchell_2021}. 
While both movements have challenged the notion of universal applicability of RAI values, our study illustrates how these challenges do indeed appear in practice and the strategies used by practitioners on the ground to resolve these challenges. 

Taken together, RAI has emerged as an umbrella term that encapsulates the above movements, encouraging critical value discourses to produce a positive impact. At the same time, departing from previous movements that focused on specific issues within AI/ML practices, RAI takes a broad institutional approach that promotes disparate AI/ML practitioners to come together, share, and enact on key values \cite{rakova2021}. Our study expands RAI discipline by surfacing on-ground challenges of diverse AI/ML practitioners attempting to engage in shared RAI responsibilities, such as collaborative value discourse and implementation. In the next section, we further unpack the notion of values, examine their roots in HCI and their current role in RAI.

\subsection{Collective Values in Responsible AI: HCI perspectives}
Science \& Technology Studies field has long examined how values are embedded in technology systems in various social and political contexts \cite{winner1986artifacts, Star_1999, latour1992missing}. In recent years, studies within HCI have built on this foundation to bring a critical lens into the development of technology. Initial studies conducted by Nissenbaum and colleagues argued against the previously held belief that technology is ``value-neutral'', showcasing how practitioners embed specific values through their deliberate design decisions \cite{flanagan2008embodying, batya97}. \textit{Value-sensitive Design} (VSD) by \citet{Friedman2013} was another step in this direction. It has been used as a reflective lens to explore technological affordances (through conceptual and empirical inquiry) as well as an action lens to create technological solutions (technological inquiry) \cite{Friedman2021, Hendry2021}. While VSD’s core philosophy remained the same, it has been extended, often in response to its criticisms \cite{dantec2009, Jafari2015a}. 

A criticism relevant to this study is practitioners’ ease in applying VSD in the industry contexts \cite{Shilton_2013}. VSD is perceived to have a relatively long turnaround time,  often requiring specialists for implementation. To overcome these challenges, Shilton proposed \textit{`value levers'}, a low-cost entry point for value-oriented conversations while building technology artifacts in the organization \cite{shilton2018, Shilton_2013}. Value levers are open-ended activities that engage participants in value-oriented discourses to develop common ground. With creative representations, value levers can transform slow and cumbersome value conversations into creative and fruitful engagements \cite{Shilton_2013}. \blue{While previous studies have applied and shaped the notion of value levers in a very specific set of contexts, such as showcasing how designers employ them in their practices \cite{Shilton_2013}, this work shows a broader utility of value levers among a diverse set of practitioners while navigating the complex space of RAI value discourse.}

\blue{Within AI/ML research, initial exploration of values were still primarily computational in nature, such as performance \cite{chaudhuri2009}, generalizability \cite{jain2008online}, and efficiency \cite{espeholt2018}. With the advent of HCI and critical studies focusing on discriminatory algorithms \cite{birhane2022A} and \textit{responsible AI}, the discussions shifted to much broader values, such as societal and ethical values, within the AI/ML field \cite{birhane2022}. These studies focused on exposing inherent biases in the models due to absence of substantive social and ethical values.} For instance, \citet{burrell2016} demonstrated how complex ML models have inherent interpretability issues stemming from a lack of transparency about how predictions were achieved. \blue{Another set of studies by \citet{Eubanks_2017} and \citet{Noble_2018} scrutinized several algorithms governing the digital infrastructures employed in our daily lives to expose discriminatory behaviors within different situations, especially in the context of fairness,  against marginalized populations.} In a similar vein, several studies have explored individual values that they felt were critical for models, such as fairness \blue{\cite{Mitchell_2021,Chouldechova2018, FriedlerSV16}}, explainability \cite{doran2017}, non-malfeasance \blue{\cite{Jobin_Ienca_Vayena_2019, Mohamed2020}}, and \blue{justice \cite{Birhane_2021, hampton2021, Kalluri_2020}}, reflecting societal norms. \blue{A common underlying factor among several of these studies was that they focused on individual values enacted in their own spaces. Recently, however, a few studies have adopted contrasting perspectives which argue that values do not exist in isolation, but often occupy overlapping and contested spaces \cite{yurrita2022, Friedman2013}.} Our study aims to provide much-needed deeper insights within this complex space by showing how practitioners engage with and prioritize multiple values in a contested space.  

Another value dimension explored is -- ``\textit{whose values should be considered while producing AI/ML algorithms} \cite{borning2012}?'' Most studies have engaged with end-users values, lending a critical lens to the deployed models and their implications on society \cite{robertson2020, kizilcec2020, wong2021, smith2020}. These studies challenged whether developing fair algorithms should be primarily a technical task without considering end-users' values \cite{Haubermann_Lutge_2022, robertson2021}. Subsequently, researchers leveraged action research (e.g., participatory approaches \cite{delgado2021}) to design toolkits, frameworks, and guidelines that accommodated end-user values in producing ML models \cite{Katell2020, krafft2021, shen2022The}.

A more relevant set of studies have recognized the importance of understanding the values that different practitioner roles embedded while producing responsible algorithms \cite{Yildirim2022How, passi2019problem, Sheuerman2020how}. Such practitioner-focused studies are critical in understanding ``how'' and ``why'' particular values are embedded in AI/ML models early on in the life cycle. However, these studies have explored particular practitioners' values in silos, leaving much to be learned about their collective value deliberations. A nascent group of studies has answered this call. For example \citet{madaio2020codesign} focused on controlled settings, where specific practitioners' values could co-design a fairness checklist as one of their RAI values. \citet{jakesh2022how} explored a broader set of values of practitioners and compared it with end-users in an experimental setting. Another relevant study by \citet{rakova2021} has explored RAI decision-making in an organizational setting, laying a roadmap to get from the current conditions to aspirational RAI practices. 

Our study contributes to this developing literature in four ways. \blue{First, within the context of Responsible AI practices, our study goes beyond the scenario-based, controlled settings, or experimental setups by focusing on natural work settings }\cite{jakesh2022how, madaio2020codesign}, \blue{which echoes the sentiment of some of the previous open-ended qualitative studies that were conducted in the organizations \cite{passi2019problem, Yildirim2022How}, but not in the context of Responsible AI practices}. Second, we focus on a diversity of stakeholder roles, who are making an explicit effort to recognize and incorporate RAI values, unlike siloed studies previously. Third, we leverage the lens of co-production \cite{jasanoff2004idiom} to study RAI values in natural work settings. Fourth, our study extends \cite{rakova2021} by explicitly unpacking the co-production challenges deeply rooted in RAI values. To this end, we answer two research questions:
\\ \textbf{(RQ-1)}: \textit{What challenges do AI/ML practitioners face when co-producing and implementing RAI values ?} 
\\ \textbf{(RQ-2)}: \textit{In response, what strategies do practitioners use to overcome challenges as they implement RAI values ?}

\subsection{Co-production as a Lens}
To answer our research questions, we employed the conceptual framework of \textit{co-production} proposed by \citet{jasanoff2004idiom}. She defined co-production as a symbiotic process in which collective knowledge and innovations produced by knowledge societies are inseparable from the social order that governs society. Jasanoff characterized knowledge societies broadly to include both state actors (e.g., governments) and non-state actors (e.g., corporations, non-profits) that have an enormous impact on the communities they serve. Studying co-production can help scholars visualize the relationship between knowledge and practice. Such a relationship offers new ways to not only understand how establishments organize or express themselves but also what they value and how they assume responsibility for their innovations.

To operationalize co-production in our study, we invoke three investigation sites, as Jasanoff proposed. The first site of exploration is the \textit{institutions} containing different structures that empower or hinder individuals to co-produce. The second site examines different types of \textit{discourse} that occur as part of the co-production activities. Solving technological problems often involve discourses producing new knowledge and linking such knowledge to practice. The last site of co-production is \textit{representations} produced both during co-production to facilitate discourses and after co-production in the form of the end-product. The three sites of the co-production framework are appropriate for understanding the current industry challenges around RAI innovation for several reasons. Technological corporations developing AI/ML innovations have a robust bi-directional relationship with their end-user communities.
Moreover, for successful RAI value implementation, practitioners need to leverage the complex structures within their organizations that are invisible to external communities. RAI value implementations occur through strategic discourses and deliberations that translate knowledge to effective execution. Lastly, in the process of RAI value deliberations, individuals co-create representations that further the implementation efforts of RAI.

%% file: 3-methods.tex
\section{Methods}
To answer our research questions, we conducted a qualitative study consisting of 23 interviews with active AI/ML practitioners from 10 different organizations that engaged in RAI practices. After receiving internal ethics approval from our organization, we conducted a three month study (April-June, 2022). In this section, we briefly talk about the recruitment methods and participant details.


\subsection{Participants Recruitment and Demographics}
To recruit AI/ML practitioners who actively think and apply RAI values in their day-to-day work, we partnered with a recruitment agency that had strong ties with different types of corporate organizations working in the AI/ML space. We provided diverse recruitment criteria to the agency based on several factors, including gender, role in the company, organization size, sector, type of AI/ML project, and their involvement in different kinds of RAI activities. Using quota sampling technique, the agency advertised and explained the purpose of our study in diverse avenues, such as social media, newsletters, mailing lists, and internal forums of different companies. For the participants that responded with interest, the agency arranged a phone call to capture their AI/ML experience, as well as their experience with different RAI values. Based on the information, we shortlisted and conducted interviews with 23 AI/ML practitioners who fit the diverse criteria mentioned above. \blue{The aforementioned factors were used to prioritize diverse participants with experience working on RAI projects within their team in different capacities. For example, while shortlisting, we excluded students working on responsible AI projects as part of their internships and included individuals who were running startup RAI consultancy firms}. 
\input{4-table}
\blue{Out of the 23 practitioners, 10 identified themselves as women. Participants comprised of product-facing roles, such as UX designers, UX researchers, program/product mangers, content \& support executives, model-focused roles, such as engineers, data scientists, and governance focused-roles, such as policy advisors and auditors.}
Out of 23 practitioners, all but one participant worked for a U.S. based organization. However, participants were geographically based in both Global North and Global South. \blue{Participants also worked in a wide variety of domains, including health, energy, social media, personal apps, finance and business among other, lending diversity to the captured experiences.}
Three participants worked for independent organizations that focused exclusively on RAI initiatives and AI governance. twelve participants had a technical background (e.g., HCI, computer-programming), four had business background, two had law background and one each specialized in journalism and ethics. For more details, please refer to \blue{Table  \ref{tab:org-participants}}.

\subsection{Procedure}
We conducted semi-structured interviews remotely via video calls. Before the start of the each session, we obtained informed consent from the participants. We also familiarized participants with the objective of the study and explicitly mentioned the voluntary nature of the research. The interviews lasted between 40 minutes and 2 hours (avg.= 65 mins.) and were conducted in English. Interviews were recorded, if participants provided consent. Our interview questions covered different co-production practices. First, in order to understand different co-production challenges (RQ-1), we asked questions about (1) how practitioners faced challenges when sharing RAI values across roles (e.g., ``\textit{Can you describe a situation when you encountered problems in sharing your values?}’’ ) and (2) how practitioners faced challenges when collaborating with different stakeholders (e.g., ``\textit{What challenges did you face in your collaboration to arrive at shared common responsible values?}’’). Second, to understand different co-production strategies (RQ-2) we asked (3) how practitioners handled conflicts (e.g., ``\textit{Can you give an example where you resisted opposing peers' values?}’’) and (4) how practitioners sought assistance to achieve the alignment in RAI values (e.g., ``\textit{What was the most common strategy you took to resolve the conflict}?’’). To invoke conversations around RAI values, we used a list of RAI values prepared by \citet{jakesh2022how} as an anchor to our conversations. After first few rounds of interviews, we revised the interview script to ask newer questions that provided deeper understanding to our research questions. We stopped our interviews once we reached a theoretical saturation within our data. We compensated participants with a 75\$ gift-voucher for participation. 

\subsection{Data collection and Analysis}
Out of 23 participants, only three denied permission to record audio. We relied on extensive notes for these users. Overall 25.5 hours of audio-recorded interviews (transcribed verbatim) and several pages of interview notes were captured. We validated accuracy of notes with the respective participants. Subsequently, we engaged in thematic analysis using the NVivo tool. We started the analysis by undertaking multiple passes of our transcribed data to understand the breadth of the interviewee’s accounts. During this stage, we also started creating memos. Subsequently, we conducted open-coding on the transcribed data while avoiding any preconceived notions, presupposed codes, or theoretical assumptions, resulting in 72 codes. We finalized our codes through several iterations of merging the overlapping codes and discarding the duplicate ones. To establish validity and to reduce bias in our coding process, all the authors were involved in prolonged engagement over multiple weeks. Important disagreements were resolved through peer-debriefing \cite{creswell2000determining}. The resultant codebook consisted of 54 codes. Example codes included, `social factors’, `prior experience’, `enablers’, `RAI pushback’.  As a final step, we used an abductive approach \cite{Timmermans_Tavory_2012a} to further map, categorize, and structure the codes under appropriate themes. To achieve this, we used three key instruments of co-production framework developed by \cite{jasanoff2004idiom}, namely, \textit{making institutions}, \textit{making discourses}, and \textit{making representations}. Examples of the resultant themes based on the co-production instruments included `value ambiguity’, `exploration rigidity', `value conflicts’, and `value lever strategies'. Based on the instruments of co-production framework, we have present our resultant findings in the next section. 

%% file: 4-table.tex
\begin{table*}
 \center
 \renewcommand\arraystretch{1.3}
 \footnotesize
 \begin{tabular}[t]{|p{0.6in}|p{1.7in}|p{0.8in}|p{1.9in}|}
 \hline
 \textbf{Participants} & 
 23 &
 \textbf{Gender}& Men: 13 , Women: 10 \\
 \hline
 \textbf{Age (years)}  & 
 Min: 30-35; \blue{Max: 55-60; Avg: 35-40} &
 \textbf{Experience (years)} & 
Min: 1; Max: 15.6; Avg: 3.94\\

 \hline
 \textbf{Company type} & 
 Product: 16, Service: 7 &
\textbf{Company Scale} & 
Small: 4, Medium: 4, Large: 15\\
 
 \hline
 \textbf{Region} & 
  Global North: 18, Global South: 5 &
 \textbf{Roles} &
 Engineer: 5, UX Designer/Researcher: 3, Product/Program Manager: 5, Senior Management (e.g., director): 3, Content \& Support: 2, Policy \& governance: 5\\

 \hline
 \textbf{ML Type} & 
 Supervised: 7, Unsupervised: 12, Reinforcement Learning:9  Deep learning: 13 &
\textbf{ML application} & 
Health: 6 , Energy: 1, Social media: 3, Personal apps: 7, Finance: 2, Business: 5, Work: 2\\
\hline
 \end{tabular}
  \caption{Practitioners' Demographic Details.}
 \label{tab:org-participants}
\end{table*}

%% file: 5-1-structures.tex
\section{Findings}
Our overall findings are divided based on different sites of exploration proposed by \citet{jasanoff2004idiom}. The first section answers RQ-1 by exploring several \textit{institutional} challenges that hinder the co-production of RAI values among practitioners (Section \ref{section-1}). The second section explores subsequent knock-on challenges that unstable institutional structures create in co-production \textit{discourses} (Section \ref{section-2}). The last section answers the RQ-2 by presenting carefully thought out \textit{representations} that overcome challenges deliberation and execution of RAI values using the concept of value levers \cite{Shilton_2013}. (Section \ref{section-3}).

\subsection{RAI Value Challenges within the Institutional Structures} \label{section-1}
Institutional structures are essential in enabling co-production of new knowledge \cite{jasanoff2004idiom}. It is these structures that facilitate relationships for deliberation, standardize democratic methods, and validate safety of new technological systems before information is disseminated into the society. We found two key institutional structures that facilitated deliberation around RAI values within AI/ML companies. These structures brought about different RAI challenges.

\bheading{Bottom-up: Burdened Vigilantes}
The First type of structures were bottom-up. Within these structures, RAI conversations developed through RAI value sharing in the lower echelons of organizations, often within AI/ML practitioners' own teams. In our interviews, eight practitioners, \blue{namely a UX researcher, designer, content designer, and program manager from two mid-size organizations and two large-size organizations experienced or initiated bottom-up practices that engaged with RAI values}. One of the enablers for such bottom-up innovation was individuals' sense of responsibility towards producing AI/ML models that did not contribute to any harm in society. A few other practitioners paid close attention to `\textit{social climate}' (e.g., LGBTQ month, hate speech incidents) to elicit particular RAI values. For instance, P08, a program manager in a large-scale technology company took responsibility for RAI practices in their team but soon started supporting team members to come together and share RAI values:

\begin{quote}
    ``\textit{
    We cater to projects that are very self-determined, very bottom-up aligned with our values and priorities within the organization \dots These are what I call responsible innovation vigilantes around the company. I also started that way but have grown into something more than that. You'll see this at the product or research team level, where somebody will speak up and say, `Hey, I want to be responsible for these RAI values, make this my job and find solutions'. So you start to see individuals in different pockets of the company popping up to do RAI stuff.}''
\end{quote}

\begin{figure*}
\centering
\includegraphics[width=\textwidth,scale=1.5]{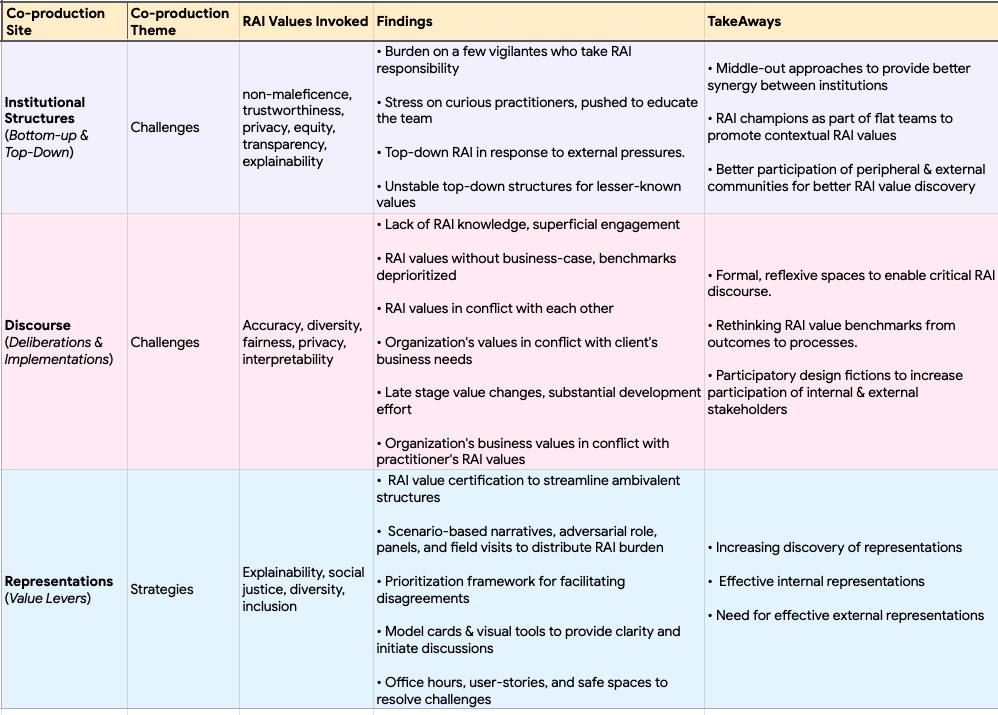}
\caption{A summary of co-production activities mapped to \citet{jasanoff2004idiom}'s co-production sites, along with the themes, RAI values invoked, and key findings and takeaways.}
\label{fig-1}
\end{figure*}

A key challenge with such bottom-up structures was that the responsibility of engaging with RAI value conversations implicitly fell on a few individual ``\textit{vigilantes}''. They had to become stalwarts of particular RAI values and take out substantial time out of their work to encourage and convince their teams to engage with RAI values. They also actively seeked out RAI programs available within and outside their organization. When such RAI programs were not available, individuals took it upon themselves to create engagement opportunities with other members within the organization. \blue{These bottom-up structures were useful in breaking the norms of ``boundary-work'' that are often set within AI and similar technical organizational work where only engineers and high-officials in the company maintain control \cite{Gieryn_1983}. It allowed non-technical roles such as user experience researchers, product managers, analysts, and content designers to a create safe space and lead the RAI efforts.} While such efforts early on in AI/ML lifecycle minimized the potential harm of their ML models or AI products, it often happened at the cost of their overworked jobs.

\bheading{Bottom-up: Burdened with Educational Efforts}
Apart from self-motivated vigilantes, the burden of RAI value exploration also fell on a few practitioners who were implicitly curious about RAI innovation. \blue{ Unlike the vigilantes, these participants were pushed to become the face of their team's RAI initiatives since there was no-one else who would. P14, a product manager working for two years at a medium size company within the energy sector shared,} 

\begin{quote}
    ``\textit{\blue{When I came in to this team, nobody really believed in it [RAI] or they really didn't think it [RAI] was important. I was personally interested so I was reading about some of these principles \dots When there was an indication of a future compliance requirement, people didn't want to take up this additional work \dots somebody had to do it.}}''
\end{quote}

Similarly P05, a technical manager leading their team on data collection for the development of knowledge graphs, revealed how they were considered ``\textit{the face of privacy}'' for the team. Therefore, P05 was expected to foster awareness and common understanding among internal stakeholders and external partners and ensure they strove towards similar RAI standards and appreciated data-hygiene practices (e.g., data cleaning and de-identification). \blue{Practitioners like P14 and P05 had to assume the responsibility of figuring out the RAI space by presenting their team's needs and asking formative questions even when their objectives around RAI were often not clear, such as which values to consider (e.g., ``\textit{privacy or transparency?}''), what certain values mean (e.g., ``\textit{what trustworthiness as an RAI value should mean to the model and the team}''), how to operationalize specific values (e.g., \textit{``How does trustworthiness apply to rule-based models? What kind of RAI values to invoke while collecting data?’’}, and how to interpret outcomes and map them on to their team's objectives.} 


\blue{Participants (n=5) shared how leading such RAI initiatives burdened their professional lives in various ways. Multiple participants reported that the RAI field was still in its infancy and taking up responsibilities in such conditions meant that their efforts were not deemed a priority or sometimes even officially recognized as AI/ML work \cite{madaio2020codesign}. Consequently, the practitioners possessed limited understanding of the direction to take to educate their team, convert their efforts into tangible outcomes, and effectively align their educational outcomes to the team's objectives. P13, an RAI enthusiast and an engineer at a large-scale social media company shared how their RAI effort seemed like an endless effort, \textit{``At this point, I probably know more about what things (RAI values) we don't want in it (model) than what we do want in it \dots It's like I am just learning and figuring out what's missing as I take every step \dots It is unclear which [RAI] direction will benefit the the team.''}} Moreover, the burden of educating multiple team members was on the shoulders of a very few practitioners tantamounting to substantial pressure.

\blue{\citet{Metcalf_2019} in their paper around technology ethics put forward the term `\textit{ethic owners}'. This role shares similarity with the bottom-up vigilantes and the educators as they both are motivated and self-aware practitioners, invested in foregrounding human values by providing awareness and active assistance while institutionalizing the processes. However, Metcalf's ethic owners' responsibilities were clearly defined. Their tasks of working with teams or higher management were perceived as visible, prioritized work for which they would be credited for career growth or otherwise.  While bottom-up informal roles in our research performed similar set of tasks, their efforts were seen as tangential,  `administrative', and underappreciated. It is not just that there was an additional burden, but even the outcomes of taking on this additional burden for bottom-up informal roles were dissimilar to the ethic owners. Taking up informal RAI work was more problematic when the requirements in the later stages of ML were unprompted, compelling practitioners to focus on these efforts at expense of their own work.} 

In our findings, one form of the need came as academic criticism or critique around particular values that were seen concerning a particular product (e.g., ``\textit{what steps are you taking to ensure that your model is equitable?’’}). Another form of need came from end-users' behavior who experienced the models through a particular product. P20, a user experience researcher working with deep learning models in finance, shared how user feedback brought about new RAI needs that became their responsibility:

\begin{quote}
    ``\textit{Once the users use our product and we see the feedback, it makes us realize, `oh, people are sometimes using this feature in an unintended way that might in turn impact the way we are going about certain values, say transparency' \dots. Initially we were like, `We should strive for transparency by adding a lot of explanations around how our model gave a particular output'. Later we realized too many explanations [for transparency] fostered inappropriate trust over the feature\dots UXR represents user needs so its on me to update the team on the issues and suggest improvements.}
\end{quote}

\blue{A few practitioners (n=2) also mentioned how the constant juggling between their own role-based work and the unpredictability of the RAI work pushed them to give-up the RAI responsibilities all-together.}

\bheading{Top-down: Rigidity in Open-discovery}
While the burden of ownership and execution of RAI values in bottom-up structures were on small group of individuals, they had the flexibility to choose RAI values that were contextual and mattered to their team's ML models or projects. On the contrary, we found that top-down institutional structures limited the teams' engagement to ``\textit{key}'' RAI values that impacted organization's core business values. For instance, P15's company had \textit{trust} as a key value baked into their business, requiring P15 to focus on RAI values that directly reduced specific model's biases, thereby increasing the company's trust among their users. Consequently, several RAI practitioners had to skip RAI value exploration and sharing. Instead they directly implemented predetermined RAI values by the management just before the deployment. P06, an engineer at a large tech company working in conversational analysis models, described this lack of choice:

\begin{quote}
``\textit{To be honest, I imagine lots of the conversations, around the sort of values that need to go into the model, happened above my pay grade. By the time the project landed on my desk to execute, the ethics of it was cleared and we had specific values that we were implementing.}''
\end{quote}

Public-oriented legal issues and ethical failures, especially when launching innovative models (e.g., transformer networks), also determined the RAI values that were prioritized and the subsequent formal RAI structures that were established by the organizations. P19, a policy director at a RAI consultation firm facilitating such structures, shared how such impromptu structures were quite common in response to ever-changing laws around AI governance:

\begin{quote}
    ``\textit{Even if you're conservative, the current climate is such that it's going to be a year or two max from now, where you will start to have an established, robust regulatory regime for several of these (RAI) issues. So a good way to be prepared is to create the [RAI] programs in whatever capacity that enables companies to comply with the new regulations, even if they are changing because if you have companies doing Responsible AI programs, it eventually gets compliance and executive buy-in. }''
\end{quote}

\blue{Instead of investing in RAI structures to comply with different regulations in Tech Ethics such as codes of ethics, statements of principle, checklists, and ethics training as meaningful, organizations perceive them as instruments of risk that they have to mitigate \cite{Metcalf_2019}. In line with the previous literature \cite{Green_2021, Bietti2020}, our findings indicate that practitioners often find false solace in such structures as they run the risk of being superficial and relatively ineffective in making structures and practices accountable and effective in their organizations. However, adding nuance to this argument in the case of RAI practices, we found that practitioners more broadly devoted time and energy to follow established and prioritized values (e.g., trust or privacy) due to directed and concerted focus. It allowed for organization-wide impact since the ``buy-in'' already existed \cite{madaio2020codesign}}.

\bheading{Top-down: Under-developed Centralized Support}
However, in the case of less clearly defined values (e.g., non-maleficence or safety) we observed a limited scope for nuance and despite best efforts, the centralized concerted direction does not always pan out as intended. 
Further, while laws continue to evolve in this space, participants felt that pre-mediated RAI values might not longitudinally satisfy the growing complexity of the ML models being implemented (e.g., multimodal models). Hence, while it might seem that setting up a centralized top-down approach might be efficient, the current execution leaves much to be desired. In fact, based on data from over half the participants, we found that five top-down structured companies integrated lesser known RAI values into their workflows in multiple ways without establishing a centralized workflow. Those who did establish centralize workflows created consulting teams to advise on RAI Practices (similar to ethic owners \citet{Metcalf_2019}).

However, these top-down centralized RAI consulting teams were not always set up to succeed. As is the nature of consulting, people did not always know the point-of-contact or when and how to reach out. The consulting teams needed to also consider opportunities to advertise about themselves and engagement mechanisms,\blue{ which was difficult due to the lack of context and nuance around the teams' projects. Consequently, it was difficult for such teams to generate organic interest, unless the teams were already aware of their RAI requirements and knew a point of contact.} P10, a manager who facilitated one such top-down RAI program in a large-scale technology company for AI/ML teams, described lack of fixed ways in which teams engaged with them on RAI values, making it a difficult engagement: 

\begin{quote}
``\textit{We have a bunch of internal Web pages that point you in all different types of directions. We don't have a singular voice that the individuals can speak with \dots. Its currently hodgepodge. Some teams come to us willingly. They had already thought about some harms that could occur. They say, `Here's our list of harms, here’s some ideas on what we want to do’ They'd already done pre-work and are looking for some feedback. Other teams come to us because they've been told to. \dots They haven’t thought much about RAI and need longer conversations \dots Other teams were told to go track down an individual or team because they are doing ML stuff that will require RAI assistance, but they don't know about us}’’
\end{quote}


%% file: 5-2-discourse.tex
\subsection{Challenges within RAI value Discourses} \label{section-2}
Fruitful co-production requires well-established institutional structures that can empower stakeholders to engage in stable democratic \textit{discourses} with an aim of knowledge production \cite{jasanoff2004idiom}. In the previous section, we uncovered different structural challenges at the institutional level that contributed to knock-on effects, creating further challenges for practitioners during the co-production and implementation of RAI values.


\bheading{Discourse: Insufficient RAI Knowledge} 
A key challenge that many practitioners experienced in co-producing RAI values in team was the difficulty in engaging deeply with new and unfamiliar RAI values deemed important by the team members. P07, a policy advisor in a large technology company, who regularly interacted with those who implemented RAI values, described the ``\textit{superficial engagement}'' with values as an act of ``\textit{ineffective moralizing}'', wherein practitioners struggled to develop deeper interpretations of the team's shared values and contextualize them in relation to the ML models they were developing.

P07 mentioned several key critical thinking questions that AI/ML practitioners did not deliberate within their teams, such as ``\textit{Is this RAI value applicable to our product?}'', ``\textit{how does this value translate in diverse use cases?}'', or ``\textit{should this value be enabled through the product?}'' The need for deeper engagement becomes particularly important in a high-stakes situation, such as healthcare, where certain conditions have an unequal impact on particular demographics. P12 experienced this complexity while overseeing the development of a ML model focused on health recommendations:

\begin{quote}
    ``\textit{So a lot of models are geared towards ensuring that we have we are predictive about a health event and that almost always depends on different clinical conditions. For example, certain ethnic groups can have more proclivity to certain health risks. So, if your model is learning correctly, it should make positive outcomes for this group more than the other groups. Now, if you blindly apply RAI values without thinking deeply about the context, it might seem that the model is biased against this group when in reality these group of people are just more likely to have this condition, which is a correct conclusion, not a biased one.}''
\end{quote}

Such deeper analysis requires hands-on practice and \blue{contextual training in the field and formal RAI education. In our findings, top-down structures were only effective to fill the gap for \textit{key} values that aligned with the company's vision, leaving a much needed requirement for contextual, high-quality RAI education for more emergent RAI values that could be modularized for specific teams. P02, a content designer for a large health technology company shared how this gap looked like for their team that was designing content for a machine translation team},

\begin{quote}
    \textit{``\blue{One thing that would have been beneficial is if I or my team could somehow get more insights on how to think about trustworthiness in the context of the content produced by our machine translation model and probably evaluate it \dots Often time, I just go to someone who is known to have done some work in this [RAI] and say, `Hey, we want to design and publish the content for the model like in a month from now, what is bare minimum we could do from [RAI] principles point of view?'\dots Sometimes it's never-ending because they say I have not thought about this at all and that it is going to take a month or maybe much more longer to get these principles implemented.}''}
\end{quote}

\blue{Participants like P02 had no alternative but to reach out to their bottom-up structures to seek assistance, discuss, and reduce gaps in their RAI knowledge. On occasions, such avenues of discussion were non-conclusive. Prior literature in AI/ML and organization studies have shown how such unequal dependence on bottom-up structures over top-down in deliberation can contribute to tensions, and in turn propose an ``\textit{open, federated system}'' linking different actors, resources, and institutions to provide a community based support \cite{rakova2021, Scott_Davis_2016}.}

\bheading{Discourse: Deprioritized Unfamiliar \& Abstract Values}
Naturally, practitioners tried to solve the superficial engagement problem by de-prioritizing values that they found unfamiliar. In our study, most practitioners (n=18) said that they were familiar and comfortable talking about RAI values like \textit{privacy} and \textit{security} as they were already ``\textit{established}'' and had ``\textit{matured over time}''. They sharply contrasted this perceived familiarity with other RAI values like \textit{explainability} and \textit{robustness}. \blue{The familiar values were well backed with stable top-down structures and dedicated teams, such as compliance departments and dedicate RAI personnel}
, making it easy for practitioners to develop mental models of deeper engagement. P20 shared their experience in this regard in their organization:

\begin{quote}
    ``\textit{The ideal situation would be like, `Oh, I have certain RAI principles that I want to make sure our product has or addresses'. In reality not all the principles are thought out the same way and applied in the first go. It usually happens in layers. First and foremost, people will look at privacy because that's super established, which means everyone knows about it, they already have done probably some work around it, so its easy to implement. And then after that, they're like, `Okay, now let's look at fairness or explainability' \dots We usually have to be quick with turnaround like one or two months. Its nice to bring up values that are new but naturally they also require familiarizing and implementation effort within the team and people see that}''
\end{quote}

Other practitioners (n=3) also followed a similar de-prioritization process for RAI values that they felt were abstract and did not have a measurement baseline (benchmarks) as opposed RAI values that could be easily measured quantitatively against a baseline. An example observed in this category was the contrast between RAI values like interpretability, which had concrete implementation techniques and measurements (e.g., LIME) and non-maleficence, which did not have a clear implementation technique or measurements. \blue{Similarly, practitioners (n=2) who went out of their way to understand and suggest new interpretability techniques for model debugging techniques (e.g., Integrated Gradients, SHAP) found it disempowering when their team members often negotiated for easier and computationally cheaper values like accuracy (e.g., P/E ratio) for implementation}. 


\bheading{Discourse: Value Interpretation Tensions}
Even in situations, when different practitioners took a similar balanced approach to prioritization, tensions emerged as different roles interpreted and contextualized the RAI values differently during the value deliberations. We found these tensions occurring \textit{within practitioners} when different practitioners defined RAI values (e.g., equity) and mapped them to RAI features and metric (e.g., skin tone) differently. P18, a senior data scientist leading an AI team in a non-profit institute, shared one such similar tension among their team members working on the same project,

\begin{quote}
    ``\textit{Probably half of my colleagues do believe that there is a cultural, and historical set of RAI values that can be applied to all the products organization wide. Other half are vehemently opposed to that concept and say that [RAI] values are always model and project dependent. So if you are talking about our long-term goal to establish a set of RAI principles, whose perspectives should be considered?\dots This is an uneasy space that needs careful navigation.}''
\end{quote}

While deliberations might occur between team-members, they might occur within a practitioner, or between the team and the end-consumers of the product/service. Latter two usually surfaced with user-facing roles, e.g., Product Managers or UX Researchers. These roles have the responsibility to understand, internalize, and embody end-user values in addition to their own values. Overall, we found that practitioners in these roles had to invest more effort to tease out their own values from that of the end-users. P04 was a user experience researcher working on interfacing a large language model for natural conversations with users. 
While P04 was interested in eliciting better insights from the model's behavior issues (i.e. interpretability \cite{doran2017}), end-users were interested in a simplified understanding of the model's opaque behavior (i.e. comprehensibility \cite{doran2017}). A UX Researcher is, however, expected to be the voice of the user in the process. Consequently, they had the constant burden to elicit both sets of values appropriately.  

Another set of tensions also occurred between \textit{practitioners and end-users}. P22, an analyst in a financial firm, described how ML practitioners perceived RAI values to be mutable and negotiable, allowing them to implement a particular RAI value in stages instead of all at once. Such a process allowed P22 (and three other participants who reported similar narratives) to build the required experience and embed the value in the ML model or AI product. However, end-users expected these embedded RAI values as absolute and non-negotiable and not on a ``\textit{sliding spectrum}'' because they are ``\textit{they are often the list of ignored rights}'', leading to practitioner-user RAI tensions. 

\blue{Our findings show that tensions that arose from non-uniform RAI value knowledge and subsequent disparate value interpretations were unproductive and a significant obstacle in the overall co-production process of RAI values. This can be attributed to a nascent RAI field that has given rise to new forms of values (e.g., explainability, interpretability) whose definitions and contexts which keep changing. This is in contrast with prior value studies in HCI studies where the tensions and conflicts around relatively more established values (e.g., privacy) do not occur until the implementation stage \cite{sarah2020, Eriksson2022, Friedman_Hendry_2019}. 
Our findings show that the majority of value tensions occur much earlier in the value interpretation stage, often contributing to the abandonment of the value discussions altogether.}

\bheading{Implementation: RAI Values and Conflicts within}
Implementation of RAI values was also not a straight forward process as implementing certain RAI values created conflict with other RAI values. For instance, P1, an engineer working on classification models in VR environments shared how their decision to improve accuracy by excluding instances of objects with sensitive cultural meanings (e.g., objects with LGBTQ references) also had direct repercussions on the diversity and fairness of the model. Implementing RAI values also created 
cascading dependencies on the inclusion of other RAI values. For instance, P16, a program manager working as an RAI facilitator for a big tech company, shared the issues team members experienced around cascading RAI values:

\begin{quote}
   ``\textit{One common issue I see is with teams that are dealing with model Fairness issues. Most often the solution for them is to improve their datasets or sometimes even collect new forms of demographic data to retrain their model and that opens up another rabbit hole around privacy that the team now has to navigate through and ensure that their data adhere to our privacy standards. More often than not, teams don't even realize they are creating a new issue while trying to solve their existing problem.} ''
\end{quote}

Implementation challenges also occurred when organization' business values were in tension with those of external clients. In such cases, the team's commitment to engage with RAI was at odds with clients' business priorities. P02, a technical program manager for a service company that developed ML models for clients in the energy sector, had a similar issue when their team was building a model for street light automation. After P02's team received the data and started developing the model, they pushed for the value of safety. However, it was at odds with the company's value of efficiency,

\begin{quote}
    ``\textit{We should prioritize model optimization in those areas where there are higher crime rates \dots we don't want blackouts, right? \dots Their argument was if there was a very high crime rate, such areas will also have high rate of purposefully damaging the lighting infrastructure. Prioritizing service to such areas will only create high amounts of backlogs as people will just vandalize it again. \dots So they just have different priorities. After that, our team just stopped following it up as it went into the backlog. }''
\end{quote}

P02's team gave up RAI value deliberation and implementation  altogether after their clients either deprioritized their multiple attempts to make them RAI ready or took an extremely long time to approve their requests. 

\bheading{Implementation: Unexpected Late-stage Value Changes}
Another challenge practitioners faced was encountering new RAI values during late stages of implementation. These values were not initially shortlisted. Instead, they were brought out later and sometimes championed by a very vocal practitioner, who felt deeply about it.
Such late-stage RAI values also became a part of the discussion when practitioners in the team uncovered last-moment issues (e.g., bias) during implementation that significantly impacted the model. \blue{Several participants (n=3) shared how such late-stage RAI values decreased the productivity of their overall RAI discourse and implementation efforts, leading to a negative experience}. While such last-minute changes are not welcomed, P12, an engineer shared how it also gives an opportunity to the developers to ship a better product before any harm might have been done. This tension between potentially better outcomes and slower implementation was visible in how the implementation efforts and timelines were impacted \cite{madaio2020codesign}.

Such values also disrupted a planned implementation by taking the spotlight and pushing the team in navigating the company's non-standardized approvals, thereby significantly altering the project timeline. For example, P23, an ML engineer shared how when they received issues around fairness from other stakeholders, it meant ``\textit{substantial changes to the model from the ground-up, because most of the time, issues with fairness stem from the data}''. It meant revisiting the data and redoing data collection or further debugging to remove the issues. Moreover, when new and untested RAI values assumed prominence 
(e.g., interpretability), more time and effort was required from the practitioners during implementation. RAI facilitators are essential in easing the tension in such situations by engaging in back-and-forth conversations with the teams to reduce the effort, streamline the process, and help practitioners appreciate the eventual consequences of implementing the RAI values. 

\bheading{Implementation: Perceived Misuse of RAI values}
Lastly, we also observed tensions between individual efforts in implementing RAI values and their organization's use of such efforts for the overall business purposes. For instance, P15, research director of a large-scale technology company overseeing research in large language models, shared how he was actively supporting a few teams in his company to co-produce and embed explainability into their models. However, he also expressed his concern about how companies could misrepresent such embedded RAI values,

\begin{quote}
    ``\textit{I worry that explainable AI is largely an exercise in persuasion. `This is why you should trust our software' rather than `This is why our software is trustworthy' \dots I'm not saying everybody who does explainable AI is doing that kind of propaganda work, but it's a risk. Why do we want our AI to be explainable? Well, we'd like people to accept it and use it \dots Explainability part is ethically complicated \dots even for explainability for the practitioners the company wants it to be explainable, transparent, reliable, all those things as a means to an end. And the end is `please like our model, please buy our software'}''
\end{quote}

We found two more practitioners who raised similar concerns with other RAI values, such as privacy and trust. They were concerned that making their product ``\textit{completely responsible}'' could enable companies to market their products as nearly perfect, leading to overtrust and over-reliance. \blue{These findings align with the \textit{ethics-washing} phenomenon within the tech ethics literature which argues that companies sometimes invest in ethics teams and infrastructure, adopting the language of
ethics to minimize external controversies and superficially  engage with the proposed regulations \cite{Green_2021, Wagner_2019}. Practitioners who expressed these sentiments were quite dissatisfied with their RAI implementation work as they felt their actions were merely a ``band-aid'' solution for the organization, instead of meaningfully altering organization's culture and practices.}

%% file: 5-3-representation.tex
\subsection{Representational Strategies to Mitigate RAI Challenges} \label{section-3}
In response to the challenges mentioned in the aforementioned sections, we saw several strategies used by the practitioners to overcome the limitations in RAI value co-production. To present the strategies, we use a form of \textit{representations} called values levers \cite{Shilton_2013}, a set of activities that facilitate opportunities to share and collaborate around values. We show how different practitioners use value levers to build upon current RAI institutional structures and make their RAI co-production  manageable. In an ideal situation, value levers can also be employed in any situation of RAI co-production. For example, organizations created several formal RAI structures for practitioners to facilitate sharing and deliberation of values. These included top-down standardized guidelines, such as guidebooks (e.g., PAIR \cite{PAIR:}, HAX \cite{HAX:}) around established RAI values, bringing in experts to share their experiences around co-production (lectures), and enabling shared spaces for co-production. However, in this section, we will be looking at value levers specifically developed in response to the challenges experienced in RAI value co-production.


\bheading{Institutional Value Levers: External Expertise and Certifications to reduce Ambivalence} 
One of the ways in which organizations brought stability to their inconsistent top-down RAI institutional structures was by taking assistance of independent agencies or professionals who specialized in establishing values levers that helped streamline their existing structures. One such value lever was `\textit{Responsible AI certifications}' that were designed to bring different recognized and informal RAI co-production activities under one-roof. These programs act as facilitators between the non-technical and technical workforce by enabling co-production around RAI values to make them compliant with upcoming regulations. Participants reported that different activities were packaged into the RAI certification program, such as getting buy-in for particular RAI values, leveraging trusted partners for running impact assessments, engaging key actors in value discovery and prioritization, and implementing appropriate RAI methods. P19, policy director of one such RAI certification organization, shared how these certifications are effective in sectors, such as energy, mining, and human resources that often have a limited technology workforce. They described effort of facilitating RAI value conversations within their client teams as a key part of the certification process:

\begin{quote}
``\textit{It is important to have everybody on board for those [RAI] value conversations. So we try really hard to have all the different teams like internal or external audit, legal, business, data and AI team come together, brainstorm, discuss different [RAI] issues in specific contexts and shortlist [the RAI values], even if we just get a little bit of their time \dots everyone needs to be brought in early because we conduct a lot of activities likes audit analysis, bias testing\dots it saves time, addresses several concerns, and establish streamlined [RAI] processes. \dots For simplicity, we just package all of the different activities we do under RAI certification. \dots Some times few activities are already being executed by the organization, we just do the job of aligning them in a way that works for the organization.}''
\end{quote}

Such external expertise and certifications can provide an opportunity for open discovery, bolster existing centralized support, and identify RAI values that might otherwise be discovered at the last stages.
    
\bheading{Institutional Value Levers: Activities to Distribute RAI burden}
We also found several nascent but more focused value levers in bottom-up institutions focused on distributing the burden experienced by a few roles more widely within the team. These value levers provided opportunities for increased participation from stakeholders, especially in the starting stages by enabling them to bring-in complementary RAI values into the team. Most commonly used levers in this context included scenario-based narratives and role plays, and open-ended activities that engaged practitioners in opinion formation and sharing. Other value levers included conducting a literature review of specific RAI values and applicable cutting-edge methods, definitions, and guidelines around them to share and invoke feedback from the team. We also observed more experimental value levers that were geared towards bringing-in complementary RAI values of external stakeholders (e.g., end-users) into the team.

For example, P18, a data scientist working in a startup, hosted a panel to capture complementary perspectives around AI explainability. Visibility into how explainability was perceived differently by different community members, such as NGOs and government, contributed to a better understanding and alignment within the team to develop explainable models. In a similar example, P09, an engineer working on a reinforcement learning model in the context of healthcare for low resources in India, facilitated field visits to the end-user communities. \blue{Such exposure helped roles that were passive in sharing their values as well as roles that were thinking about new values, such as social justice, in the RAI discourse.}
Overall, these value levers (narratives, role-plays, literature reviews, panels, and field visits) focused on primarily bottom-up structures, which helped reduce pressure on specific roles and limit superficial value engagements.

\bheading{Discourse Value Levers: Facilitating Disagreements}
Moving our focus to RAI value co-production, we saw user-facing practitioners create explicit opportunities for disagreements and healthy conflicts to tackle the problem of superficial value engagement and improve the quality of their teams' deliberations. Disagreements in co-production phase allowed practitioners like UX researchers and product managers to think inclusively, capture diverse perspectives and expert knowledge, and more importantly predict future value conflicts. 
For example, P04, a UX researcher, created bottom-up adversarial prioritization framework. In the starting phases of this framework, the UX researcher pushed team members to go broad and co-produce values by wearing other practitioner's hats and invoking their RAI values. This practice allowed them to bring forward interesting disagreements between different roles that were then resolved and prioritized to achieve a small set of meaningful RAI values. P04 recalled two of the values that received maximum disagreements were diversity and inclusion. Wearing complementary user hats enabled practitioners to familiarize with these values that were otherwise unfamiliar in their own roles. Other top-down RAI programs also facilitated similar structures, explicitly providing narratives that brought out disagreements, e.g.,:
\begin{quote}
    ``\textit{Usually I will write something in the prompts that I think that the team absolutely needs to hear about but is controversial and opposing. But what I do is I put it in the voice of their own team so that it is not coming from us. It is not us scrutinizing them. That promotes interpersonal negotiation that pushes individuals to really defend their values with appropriate reasoning.}''
\end{quote}
According to P19, having such RAI system in place early also allows companies to judge its ML models benchmarks when compared to their competition. Leveraging the adversarial prioritization framework appropriately in both top-down and bottom-up structures can enable open-discovery, and surface the values and related conflicts for resolution.


\bheading{Discourse Value Levers: Model Cards \& Visual Tools to Reduce Abstractness from Values} 
We found that practitioners also created succinct representations and simplified documentation to bring much needed clarity to various RAI values and simply associated models. For instance, engineers shared documentation of model and data cards, making it easier for non-engineering and engineering roles to grasp the information. P23, a senior engineer at an AI startup looking into sustainability, shared the process:
    \begin{quote}
 `\textit{Even we have introduced this concept of a model card, wherein if a model is developed, the model card has to be filled out. So what is a model card? A model card is a series of questions that captures the basic facts about a model at a model level at an individual model level. What did you use to build a model? What was the population that was used? What is the scoring population? It is like, having all of that in a centralized standard format. Goes a long way to roll it up because the product can be very complex as well, right? With multiple players and whatnot. But having that information collected in this way benefits other roles that own the product to think about different values that are missing}'
    \end{quote}
UI/UX designers, UX researchers, and analysts also used similar documentation tools to initiate discussions and receive feedback from other practitioners in the team. P20, a UX researcher, used presentation slides containing model features to facilitate brainstorming sessions and receive feedback from other roles. \blue{They also repurposed tools and methods used in their own work to give shape to their peers' abstract values. For example, P20 reused online jam-boards containing key RAI values and user findings for affinity diagramming, enabling the team to ``\textit{categorize the findings and map them to specific RAI values}}''. Other RAI levers in this category included designing and sharing infographics and regular RAI standups where practitioners took it upon themselves to be stewards of RAI principles for the team to give updates, receive feedback and learn about team's perspectives on specific RAI values.
 
\bheading{Implementation Value Levers: Office Hours, User stories, Safe Spaces to Reduce Tensions}

A few value levers that were part of top-down RAI programs were also effective in reducing various value tensions that occurred between different practitioners  (n=2). One such program was \textit{RAI office hours} that was available for elicitation and production, but were also extremely effective for tension-resolution. A typical office hour was 30 minutes in which practitioners engaged with a relevant expert and an experienced facilitator. One key way experts solved the tensions in these sessions was by collecting and providing concrete case-study examples. For example, P21, an RAI office hour facilitator, shared an example about the use of his office hours. The practitioners were at odds with each other in implementing explainability and trustworthy features. During the office hours, P21 responded by sharing an edge case scenario where even good explanations might backfire, such as, \textit{``If a pregnant woman had a miscarriage, showing even good end-user explanations around why they are seeing infant-related content can be very problematic. Explainability should be carefully teased out based on the context in which it is applied.''}  

Another set of value levers used especially by the roles facing end-users were user stories and scenarios to influence and persuade users to change their value priorities and align with rest of the team. These levers were also used by the roles to converge on key values after engaging in healthy conflicts within the divergence phase. For example, P04, exposed different pain points and key user journeys by ``\textit{highlighting the clip of a user that is really, really amazing story that is either very painful or poignant}''. Interestingly, P04 was aware how such value levers had to be evoked carefully,

\begin{quote}
    \textit{``If that story is not representative, I'm manipulating the system. If it is representative, I'm influencing the system...I will have to be careful not operate on the side of manipulation and try to be very squarely on the side of influence. So, I do like regular checks for myself to make sure that I am operating on influence, not manipulation, in terms of the stories that I am allowing people to amplify.''}
\end{quote}

Lastly, in order to tackle several types of value conflicts in the co-production of RAI values, we found different RAI value levers that focused on improving value alignment. One key alignment strategy was to create structures and activities that aligned the team's RAI values pretty early in the process. One such activity that we saw in both practitioner's and RAI programs was providing a safe space to encourage open discussions among individuals to empathize with other members. P09 shared,

\begin{quote}
    ``\textit{One alignment strategy was open discussion with the safe space, where team members could fail, be called out and to learn from each other as we were developing values. So say someone finds the value of democratization really important, they are made to articulate what they mean by it.\dots. It is easy if there are different buckets in which they can categorize and explain because then people can easily surface all the different ways they think and prioritize values and that helps with alignment}''
\end{quote}


%% file: 6-discussion.tex
\section{Discussion and Future Work} 
Overall, our findings show that co-production of RAI values in practice is complicated by institutional structures that either support top-down decision-making by leadership or are inhabited by bottom-up practitioners exercising voluntary agency (section \ref{section-1}). In other case, multiple challenges exist when practitioners have to reconcile within their internally held values and RAI values expected from their roles; and between themselves and their team-members. \blue{Our findings also show 
that discourse around alignment and prioritization of RAI values can sometimes be unproductive, non-conclusive, and disempowering when practitioners have to implement said RAI values (section \ref{section-2}).} We observed a lack of transparency, and unequal participation within organizations; and between organizations and end-users of their products/services (section \ref{section-2}). Despite the relatively complicated lay of the land, practitioners have been pushing ahead and discovering multiple strategies on how to make progress (section \ref{section-3}). In the subsections below we will unpack these challenges, strategies and potential future work across the three sites of co-production: institutions, discourse, and representations.

\input{4-2-table}

\subsection{Envisioning balanced Institutions: Middle-out RAI Structures}
\bheading{Inequity in Burden}
According to \citet{jasanoff2001election}, strong institutions provide a stable environment for effective knowledge co-production. They can also act as safe spaces for nurturing and transforming contested ideas to effective practices leading to long-lasting impact on the immediate ecosystem. Recent scholarship by \citet{rakova2021} has put faith in an aspirational future where organizations would have deployed strong institutional frameworks for RAI issues. Our findings show that as of today, top-down structures are underdeveloped. Organizations have deployed structures that range from being reactive to external forces (e.g., compliance, public outcry) by tracking teams that implement RAI to proactively establishing structures that make teams RAI-ready (e.g., office hours). Furthermore, stable workflows have been established for a limited number of values or use-cases, restricting the number of teams that could leverage such workflows. 

Being in the midst of restrictive structures, particular practitioner roles embraced the persona of bottom-up vigilantes and self-delegated themselves to be champions of lesser known RAI values (e.g., non-maleficence and trustworthiness).
\blue{They initiate open-ended exploration for value discourses and subsequent value implementation}. However, such bottom-up structures also put burden and occupational stress on selective roles, risking the implementation success of such RAI projects. In particular, we have found that roles like UX researchers, designers, product managers, project managers, ethicists have been taking the brunt of this work. These findings build on the previous work \cite{rakova2021, madaio2020codesign}, highlighting existing inequity and subsequent individual activism being performed by some - either by volition or due to lack of alternatives.

\bheading{Enabling Equal Participation}
Going forward, there is a need for a holistic middle-out approach that seeks a balanced synergy between top-down and bottom-up structures while balancing for the challenges that each of the structures provide. For instance, organizations can work with RAI value stalwarts and champions to formalize and streamline bottom-up workflows, making it a standard practice for all teams to engage in open-ended exploration of RAI values. Such a process can enable teams to look beyond loosely applicable organization-recommended RAI values and shortlist those values that actually matter and apply to their team. To standardize the structure, organizations can leverage independent (flat) team/roles that can guide the target team through the team while giving enough room for exploration.

Organizations can also use a middle-out approach to reduce the burden and occupational stress on specific roles through several top-down activities. One such way is to facilitate structures that can lower the barrier for diverse internal stakeholders to engage in RAI value co-production, regardless of their proximity to AI products or ML models. For instance, data-workers and teams/roles that do internal testing of the models (dogfooding) can contribute to the RAI value co-production. Same structures can also enable engagement with external stakeholders, such as end-user communities, policy, experts, and governance agencies in the initial stages of value co-production. Consequently,  practitioners' chances to foresee or anticipate changing requirements could improve, especially in the later stages of AI/ML lifecycle. Better yet, this could potentially improve not just the ``user experience'' of value discourse, but also the efficiency of implementation - a goal valued by private companies. \blue{This could be a win-win situation for multiple stakeholders by helping the top-down RAI structures align with business goals. While, our research uncovered only top-down and bottom-up structures that were mutually exclusive, other structures might exist}. For example, while we envisage middle-out structures to be advantageous, future research is needed to operationalize and simulate such structures; and discover existing implementations. There might be some challenges uniquely inherent in those structures. We encourage future researchers to continue this line of enquiry

\subsection{Envisioning better Discourses: Enabling Critical RAI Value Deliberations}
\bheading{Negativity in Deliberations}
The ultimate aim of co-production discourse is to engage in competing epistemological questions. Jasanoff calls it \textit{interactional} co-production \cite[ch~8]{Jasanoff_2004c} because it deals with explicitly acknowledging and bringing conflicts between two competing orders: scientific order brought about by technological innovations and social order brought about by prevalent sociocultural practices within the community. In our findings, several underlying conflicts surfaced between scientific and social order (section \ref{section-2}). \blue{In one instance, practitioners had to choose between socially impactful but lesser explored RAI values (social order) and lesser applicable but established values with measurable scientific benchmark (scientific order). In another instance of tension an RAI value occupied competing social spaces (e.g., equity). The underlying issue was not the conflicts itself but lack of systematic structures that enabled positive conflict resolution around the conflict}. Such implicit conflicts often met with deprioritization and conversations ended unresolved.
There is an urgent need to transform such implicit conflicts to explicit positive deliberations. Organizations aiming for successful RAI co-production need to be more \textit{reflexive} \cite[ch~9]{Jasanoff_2004c}\cite{wynne1996may}, mobilize resources to create safe spaces and encourage explicit disagreements among practitioners positively, enable them to constantly question RAI values or co-production procedures that help embed them. While we saw some instances of these explicit practices in the form of value lever strategies, such instances were sporadic and localized to very few teams. 

\bheading{Acknowledging Differences Safely}
Our findings around challenges within RAI value discourse also showcase the politically charged space that RAI values occupy in an organization. In their critical piece around implementation of values in organization, \citet{borning2012} bring out a few value characteristics that are applicable to our study. First, values are not universal. Values are recognized, prioritized, and embedded differently based on a bunch of factors, such as practitioner's role, organizational priorities and business motivations which make RAI as a complex space. In our findings, roles prioritized those RAI values that were incentivized by the organizations and the AI/ML community through computational benchmarks focused on the RAI value outcomes. This is problematic as the RAI values that have computational benchmarks and have implicit organizational incentives might not map onto the RAI issues that are pertinent in the community. One way to solve this mismatch is by rethinking the definition of benchmarks from value outcome to value processes taken up by different roles (or teams) \cite{batya97}. For example, organizations can encourage teams to document their co-production journeys around lesser known RAI values that can act as RAI benchmarks.

The second issue that \citet{borning2012} bring out is \textit{whose} value interpretations should be prioritized and considered? In our findings, tensions emerged as same values were interpreted and prioritized differently by different stakeholders. End-users viewed RAI values as immutable and uncompromisable, whereas practitioners viewed them as flexible and iterable. Similar tensions were also observed between the  internal stakeholders in contextualizing the same values. While we saw a few strategic value levers, such as office hours, they were mainly targeted at the stakeholders that were within the same team. Extending this line of value levers, we propose participatory approaches that take a balanced approach for both internal and external stakeholders. In particular, we take inspiration from \textit{participatory design fictions}, a participatory approach using speculative design scenarios  \cite{vera2019, Dourish_Bell_2014}, to find alignment on polyvocal speculations around embedded values in the emergent technologies. Deliberations around the fiction narratives can be used to arrive at a common ground RAI value implementation that are both contextual and implementable.

\bheading{\blue{Shaping Institutional Structures to Improve Discourse}}
\blue{Employing co-production as a lens has also allowed us to reveal a very delicate, symbiotic relationship between the practitioners' discourses and the institutional structures. Several of the discourse challenges observed in our findings, such as issues pertaining to the lack of RAI knowledge and deprioritization of lesser known values, stemmed not only from the lack of mature top-down structures but also dependency on a singular institutional structure. The limitations of top-down structures pushed several practitioners in-turn to create informal bottom-up structures, shifting the pressure from one structure to the other. We argue that RAI-focused organizations need to seek a balance between stability (top-down structures) and flexibility (bottom-up) \cite{Weick_2006}. In addition to new middle-out \textit{internal}-facing institutional structures, RAI discourses can benefit from \textit{external}-facing institutional structures that can help such discourses be impactful. This can be achieved by bringing in diverse external collaborators, such as NGOs, social enterprises, governmental institutes, and advocacy groups. It will also help avoid the issue of co-optation\cite{green2019good} for true meaningful impact of such structures on the practices.}

\subsection{Envisioning better Representations: Value Levers}
\bheading{Value Levers Enabling Progress}
As the third site of co-production, \textit{representations} are essential manifestations of knowledge deliberations \cite[ch~3]{Jasanoff_2004c}. They provide surrogate markers for progress on value alignments or even responses to tensions between stakeholders. In our study, we saw several value levers that were deployed within organizations at different stages to tackle co-production challenges. These value levers enabled progress during knowledge deliberations (section \ref{section-3}). For instance, during deliberation, practitioners used role-specific value levers (e.g., visual tools by UX researchers \& designers) as a window into their thought process around specific RAI values. Similarly, enabling safe spaces provided opportunities of RAI value alignment among different roles. \blue{As representations, these value levers improved the transitions between internal stages and acted as vital markers to improve alignment. We call them \textit{internal representations}. Prioritization Framework employed to resolve value related conflicts across individuals was another example of internal representation.} In contrast, we also found a few externally represented value levers, such as RAI certifications, that enabled organizations to reduce ambivalence in their structures while showing their RAI readiness and compliance to the outside community. \blue{Interestingly, we uncovered limited evidence of external representations that engaged with end-user communities directly. We posit that the lack of external representations can be attributed to the deprioritized perception of the end-users' role in RAI value conversations. External representations have the potential to act as stable participatory structures that enable participation of external stakeholders, making RAI value co-production successful}. This might also be due to lack of sufficient incentives to enable such participatory structures. We hope that recent progress made by UNESCO's historic agreement \cite{unesco2020agreement} might provide the much needed push for organizations to share and learn together. 

\bheading{Value Levers Enabling Practitioners}
As we end this discussion, some readers might assume that we now have strategies to manage the multiple challenges coming our way as we deliberate and implement RAI values in products. Our research with 23 practitioners, points to the opposite - we are far from it. As one practitioner said, ``It's currently hodgepodge''. Multiple individuals and organizations are trying to surmount this incredibly complex challenge at institutional and individual level. While value levers discussed in this work were successful in helping practitioners make progress, the discovery of these value levers has at best been serendipitous. Sufficient support structures, skills training, and knowledge dissemination will be needed to enable practitioners to overcome these and unknown challenges yet. One can liken these value levers as a tool-belt for practitioners, as summarized in Table \ref{tab:strategies}. 

There is a subsequent need to design systems that can provide access to these tools in a systematic way. This would require centering research amongst practitioners who are making products. We hope that future educators, researchers, and designers will pursue this opportunity to uncover further challenges, learn from existing strategies, develop better strategies as they iterate, and create design systems to support practitioners. Further, we need to find more tools, and we need to share the tool-belt with other practitioners.

%% file: 4-2-table.tex
\begin{table*}
\center
 \renewcommand\arraystretch{1.3}
 \footnotesize
\begin{tabular}[t]{|p{0.2in}|p{2.7in}|p{2.7in}|}
\hline
 \textbf{Sn.No} &\textbf{Challenges} & \textbf{Strategies}  \\ 
 \hline
 1. & Complexity of RAI values, unstable centralized structures & 
 RAI value certification to streamline ambivalent structures\\  
\hline 
2. & Burden on a few vigilantes who have taken on responsibility & Scenario-based narratives, adversarial role, panels, and field visits to distribute RAI burden \\
& Stress on a few curious practitioners to educate the team & \\
\hline
3. & Top-down RAI in response to external pressures. Lesser RAI Autonomy &
RAI value certification to streamline ambivalent structures \\
\hline
4. &  Lack of RAI knowledge, superficial engagement & 
Model cards \& visual tools to provide clarity and initiate discussions \\
& RAI values without business case, benchmarks deprioritized & \\
\hline
5. & Certain RAI values in conflict with other values & 
 \\

& Organization values in conflict with client's business needs & Prioritization framework for facilitating disagreements
 \\

& Late stage value changes, substantial development effort & Office hours for conflict resolution, user-stories, \& safe spaces to resolve challenges\\

& Practitioner's RAI values in conflict with organization's business values & \\

\hline

\end{tabular}
  \caption{Summary of challenges mapped against strategies.}
 \label{tab:strategies}
\center
\end{table*}

%% file: 7-conclusion.tex
\section{Limitations \& Conclusions}
Based on qualitative research with 23 AI/ML practitioners, we discovered several challenges in RAI value co-production. Due to their roles and individual value system, practitioners were overburdened with upholding RAI values leading to inequitable workload. Further, we found that implementing RAI values on-the-ground is challenging as sometimes these values conflict within, and with those of team-members. Owing to the nascent stage of RAI values, current institutional structures are learning to adapt. Practitioners are also adapting by discovering strategies serendipitously to overcome the challenges. However, more support is needed by educators to educate RAI/RAI values, researchers to unpack further challenges/strategies, and for community to focus on aiding AI/ML practitioners as we collectively find a common-ground. 
Our work also has several limitations. First, our overall study is shaped by our prior HCI research experience in responsible AI. While all the participants had rich experience working with ethics in the context of social good in both global north and global south contexts, we acknowledge that our perspectives around responsible AI and ethical values in AI might be constrained. Second, our insights are also limited by the overall methodological limitations of qualitative enquiry, such as sample size of 23 participants across 10 organizations. Future work is required to generalize our insights across different organizational sectors and ML models using mixed-methods approach. 

\balance